\documentclass[10pt,twocolumn,letterpaper]{article}
\usepackage{cvpr}
\usepackage{times}
\usepackage{epsfig}
\usepackage{graphicx}
\usepackage{amsmath}
\usepackage{amssymb}
\usepackage{subfigure}
\usepackage{booktabs}
\usepackage{tabularx}
\usepackage{bm}
\usepackage{pifont}

\usepackage[pagebackref=true,breaklinks=true,letterpaper=true,colorlinks,bookmarks=false]{hyperref}

\cvprfinalcopy 

\def\cvprPaperID{2024} 

\begin{document}

\title{DeepDeblur: Fast one-step blurry face images restoration}

\author{Lingxiao Wang, Yali Li, Shengjin Wang\\
Tsinghua Unversity\\
{\tt\small wlx16@mails.tsinghua.edu.cn, liyali@ocrserv.ee.tsinghua.edu.cn, wgsgj@tsinghua.edu.cn}
}

\maketitle

\begin{abstract}
    We propose a very fast and effective one-step restoring method for 
    blurry face images. In the last decades, many blind deblurring algorithms have been proposed
    to restore latent sharp images. However, these algorithms run slowly
    because of involving two steps: kernel estimation and following non-blind 
    deconvolution or latent image estimation. Also they cannot handle face images
    in small size.
    Our proposed method restores sharp face images directly in one step using
    Convolutional Neural Network. Unlike previous deep learning involved methods
    that can only handle a single blur kernel at one time, our network is trained
    on totally
    random and numerous training sample pairs to deal with the variances due to
    different blur kernels in practice.
    A smoothness regularization as well as a facial regularization are added to 
    keep facial identity information which is
    the key to face image applications.
    Comprehensive experiments demonstrate that our proposed method can handle
    various blur kenels and achieve state-of-the-art results for small size
    blurry face images restoration. Moreover, the proposed method shows significant 
    improvement in face recognition accuracy along with increasing running speed by more
    than 100 times.
\end{abstract}


\section{Introduction}

Single image deblurring is an ill-posed task which is normally considered
as blind deconvolution process
where both sharp image and blur kernel are unknown. 
Although many blind debluring progresses have been achieved recently \cite{Lai2016A,Levin2011Efficient,patchdeblur_iccp2013,Cho2009Fast}
, applications like face 
recognition or identification still encouter troubles when face 
images are blurry due to camera shake or fast move.
Many state-of-the-art
algorithms \cite{patchdeblur_iccp2013,Fergus2006Removing,Chakrabarti2016A,Shan2008High,Levin2011Efficient,Michaeli2014Blind} 
decompose deblurring into
two stages of kernel estimation and sharp latent image estimation or non-blind
deconvolution. Methods in \cite{Levin2011Efficient,Cho2009Fast,Li2002A,Pan2014Deblurring}
estimate latent sharp images and its corresponding blur kernels iteratively.
They use salient edges and strong prior for kernel estimation of natural images.
Levin \etal \cite{Levin2011Efficient} proposes a simpler $MAP_k$ algorithm using parse derivative
prior to avoid artifacts based on Fergus \etal \cite{Fergus2006Removing}.
Whyte \etal \cite{Whyte2012Non} steps forward to discuss non-uniform blur kernel across the image.
Pan \etal \cite{Pan2014Deblurring} observes properties of text image priors and optimize
deblurring.
Those two-step method rely on blur kernel estimation and assume that kernels have no
errors when doing non-blind deconvolution. However, Shan \etal \cite{Shan2008High} proves that 
even small kernel errors or image noise can lead to significant artifacts.
Those existing methods can only handel natural images in large size while they fail 
to deblur face images which are in very small size and have little distinct sharp edges.
Also, a common drawback of these methods is slow running speed which is ineffective in
practice.
\begin{figure}[t]
    \begin{center}
    \includegraphics[width=1\linewidth]{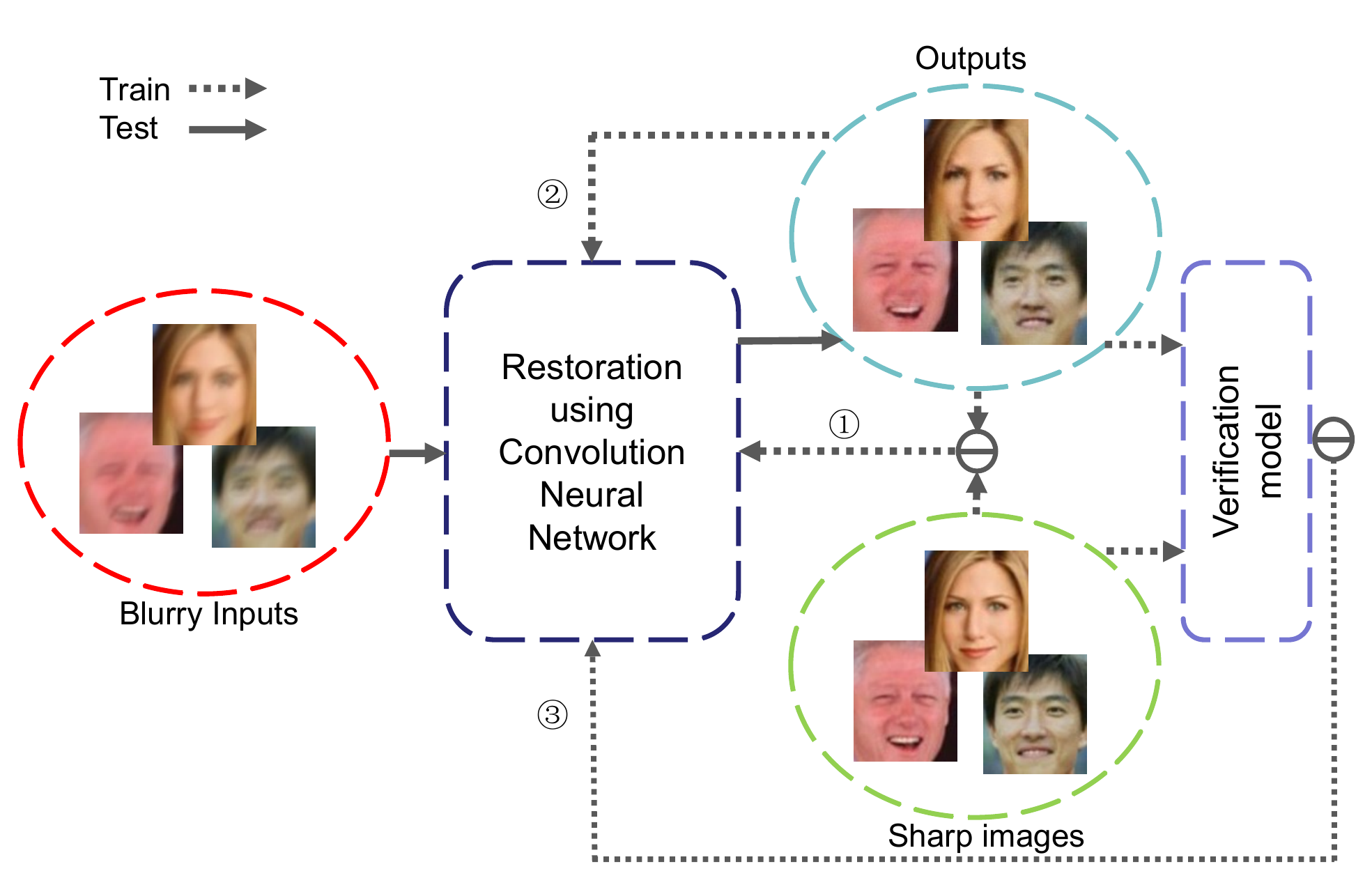}
    \end{center}
        \caption{Proposed one-step restoring method. In test phase, sharp face images are restored
        directly from blurry input. Three types regularization are added in the training phase:
        \ding{172}: L2 regularization, \ding{173}: smoothness regularization,
        \ding{174}: facial structure regularization.
        Facial identities are preserved as well as image
        qualities are improved.
        }
    \label{fig:pipeline}
    \end{figure}
\par
Deep learning has been recently widely used in image processing problems for
its strong representative ability and fast running speed
\cite{Son2017Fast,Dong2014Learning,Harmeling2012Image,Schuler2016Learning}. However,
training deep neural network with image pairs for image deblurring is very difficult
because different blur kernels lead to large variance \cite{Son2017Fast}. 
Xu \etal \cite{Xu2014Deep} proposes a non-blind deblurring method based on 
Convolutional Neural Network using large
1D kernels but it needs to be trained specifically for different kernels.
Sun \etal \cite{Sun2015Learning} trains a classification CNN network with 70 fixed blur kernels
so it cannot make a generalization. 
Son \etal \cite{Son2017Fast} applys Wiener filter to non-blind deconvolution and trains a 
residual network to remove artifacts and noise amplified by Wiener fitering.
In \cite{Chakrabarti2016A}, a neural network is proposed to predict the complex Fourier 
coefficients of a deconvolution filter to be applied to the input patch. Because it
need many overlapping patches of inputs, it cannot deblur images in small size.
Schuler \etal \cite{Schuler2016Learning} designs an iterative 
model for blind deblurring and it uses MLP structure for feature extraction module of
each iteration. This model still consists of kernel estimation and image estimation
modules and its performance is not comparable to the state-of-the-art for big blur
kernels.

In this paper, we propose a one-step restoring method using CNN
for blurry face images. Schematic of our method is shown in Figure~\ref{fig:pipeline}.
A multi-scale deep residual network is designed to handle various
blur kernels and trained on random generated training sample pairs. Our 
restoration model improves image quality as well as facial identity of each face image
by adding smoothness and facial regularizations. Compared with previous
debluring methods, the proposed algorithm is simple with one direct step and requires no 
prior or post-processing which are inevitably used in \cite{Levin2011Efficient,Fergus2006Removing,Son2017Fast}.
Moreover, experiments show that our method can restore small blurry face images  while
existing algorithms may fail or collapses. Implenting deep neural network model is very fast 
using GPUs, and our model runs more than 100 times faster than previous
debluring methods. Meanwhile, it achieves comparable results in PSNR and verification
accuracy of restored face images.

\begin{figure}[t]
    \begin{center}
    \includegraphics[scale=0.4]{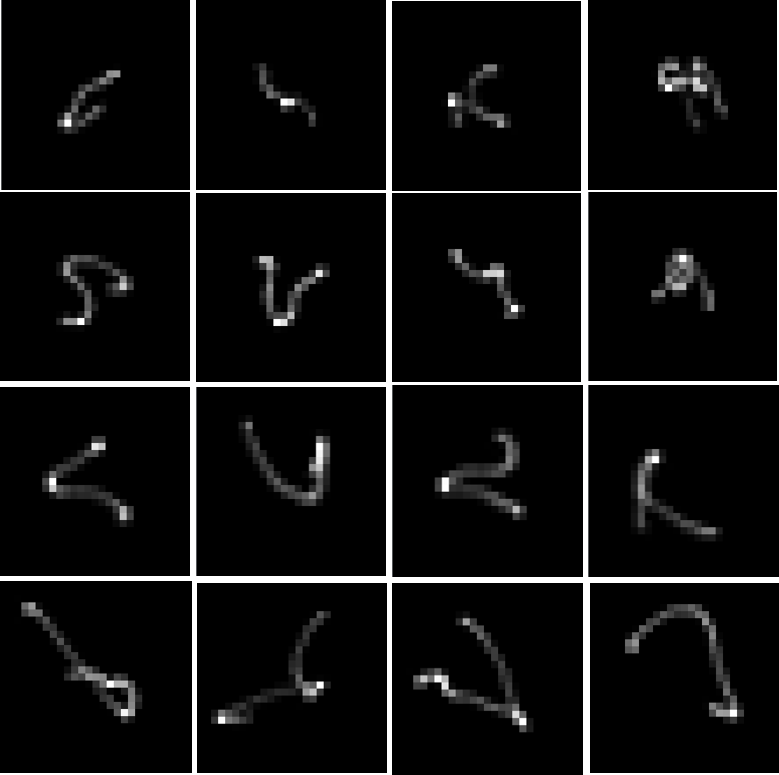}
    \end{center}
        \caption{Generated kernel samples using described gaussian process. Kernels in each
        line have different lengths. We generated kernels from valid size 8 to 20 which is
        reasonable for face images in size about 100 pixels. }
    \label{fig:kernel}
    \end{figure}

\begin{figure*}[t]
\begin{center}
\includegraphics[width=0.9\linewidth]{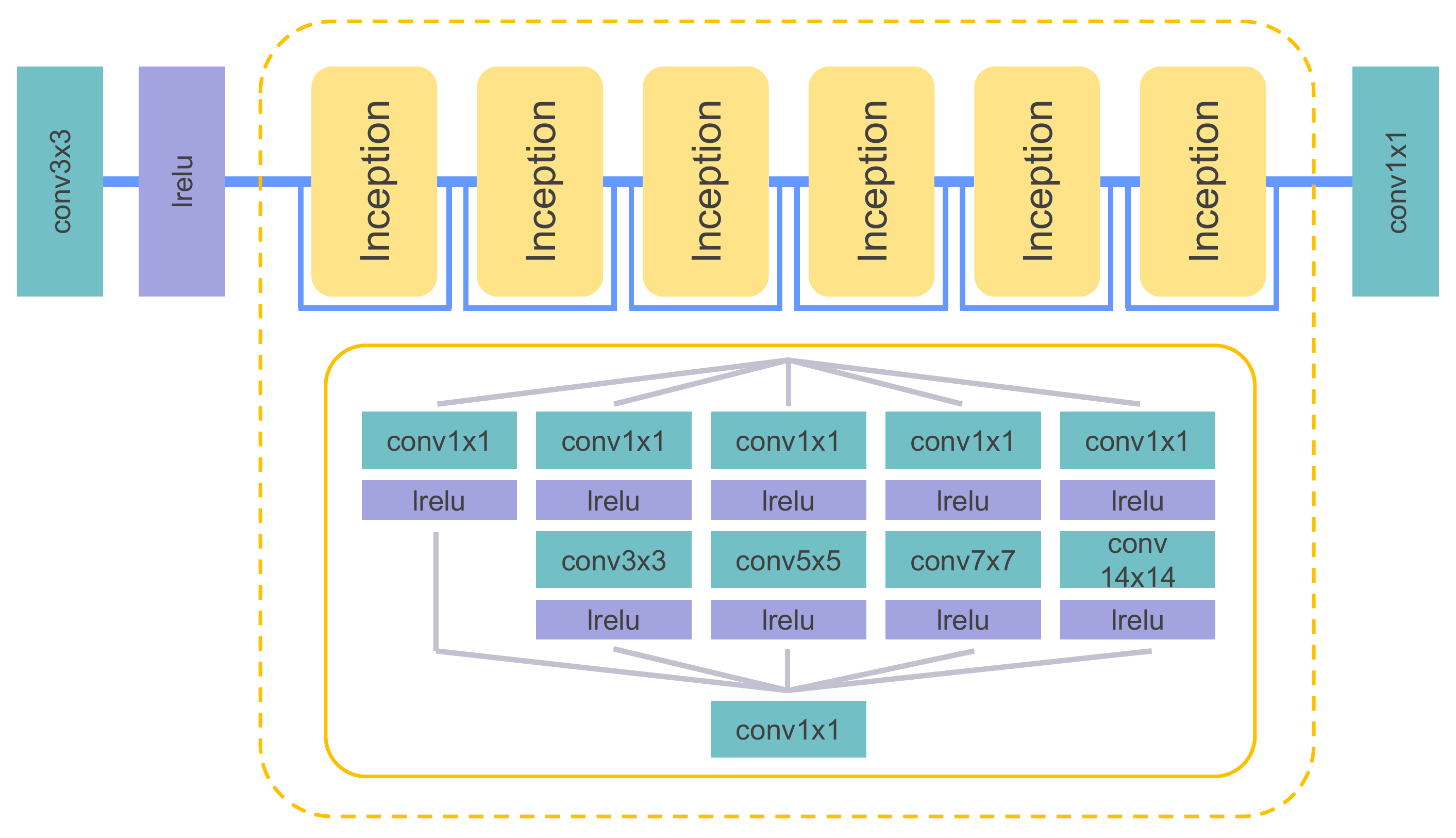}
\end{center}
    \caption{Architecture of our restoring network. Inception modules are connected in
    residual way. Input and output are both RGB images.}
\label{fig:CNN}
\end{figure*}

\section{One-step Blurry Face Images Restoration}
Single image blurring is normally modeled as a convolution process between 
an unknown latent sharp image $x$ and a blur kernel $k$ with an additional noise,
\begin{equation}
    y = x * k + n
\end{equation}
where $y$ is the blurred image and $k$ is usually unknown in practice. Correspondingly,
deblurring processing can be seen as solving a blind deconvolution problem which is 
obviously ill-posed. 
Given a blurred image $y$, $x$ and $k$ are usually estimated by solving
\begin{equation}
    \mathop{\arg\min}_{x,k}\|x*k - y\| + \beta_{x,k}
\end{equation}
Where $\beta_{x,k}$ is added as a noise term. There are usually two steps in previous works, blur 
kernel estimation based on priors and a following non-blind deconvolution. These two-step methods 
strongly rely on accuracy of estimated kernel. 
The size of a blur kernel is comparatively small to the original sharp image. Thus, a tiny variation
or error will be amplified during deconvolution and lead to artifacts. Besides,
it's always difficult for two-step methods to handle when it comes to non-uniform blurring circumstances.
\par
We find that blurred images often cause a failure of a face verification model
(see section~\ref{Quantitative}). Restoration of 
a blurred single face image also plays an important role in safety and security field.
Although many methods try to solve the general deblurring problem whose targets doesn't have 
specific content and objects, restoring sharp face images is still a huge challenge. Unlike general
deblurring datasets which have a typical height/width from 1000 to 2000 pixels, face images are very 
small. In addition, sharp edges are absent in face images, which dramaticly increase 
difficulty of the deblur process compared to natural images deblurring. However, 
from face verification tasks it has been found that
human faces have discriminative representation features which can be useful in 
restoration. 
\par 
Thus, we propose a one-step solution from blurred to deblurred face images using Convolutional 
Neural Network. The basic training of network is written as 
\begin{equation}
    \mathop{\arg\min}_{F}\|F(y) - x\|
\end{equation}
where nonlinear operation $F$ denotes neural network. There is no explicitly blur kernel
estimation in our one-step method. Also, our CNN can automatically sort and combine features that extracted
by itself from original blurred face images. 
\par
In this way, our one-step CNN method overcome influences of kernel estimation errors without
priors of blur kernel and run very fast during test. Restoration network also benefit the 
face verification which is essential in applications.

\section{CNN Based Restoration}
Below, we describe our one-step restoring method which are illustrated 
in Figure~\ref{fig:pipeline}. We introduce novel regularizations to model training
process. Samples sythesizing and network architecture are also included in this section.

\subsection{Generating Blur Kernels for CNN Training}\label{kernel_gen}
Although deblurring CNN needs no priors of blur kernel, the key to training CNN is 
sufficient training samples with different kinds of motion blurs as many as possible.
Blur degradation often comes from relative motion between camera and objects in scene.  
There are 6 degrees of freedom (3 translations and 3 rotations) in camera motion and 
they are projected in 2D image plane \cite{Lai2016A}. So we take the complexity of 
real blur
into consideration and sythesize training kernels as realistic as possible.
To avoid overfitting because of limited kernel types, kernels are synthetized as 
described in \cite{Schuler2016Learning}. It assumes that both coordinates of a 2D 
kernel follow a gaussian process with kernel function $k(t_1,t_2)$,
\begin{multline}
    k(t_1,t_2) = \sigma_f^2(1 + \frac{\sqrt{5}|t_1 - t_2|}{l} + \frac{5(t_1 - t_2)^2}{3l^2})\\
    \exp(-\frac{\sqrt{5}|t_1 - t_2|}{l})
\end{multline}
where $\sigma_f^2$ and $l$ are set as \cite{Schuler2016Learning}. Here, we set the 
sampling margin to $0.01$  and control sampling length $N$ to
obtain infinite motion kernels. 
We also randomize valid kernel size in order to stimulate different real blurs 
scales. Finally, kernels are normalized and scaled to 
one fixed size with realistic shapes show in Figure~\ref{fig:kernel}.
\par
Our deblur model is trained with pairs of a sharp face images and the corresponding 
blurry images
generated online by kernels described above. The variable kernel shapes ensure 
generalization ability of our model and avoid overfitting in finite 
datasets. 
Our model breaks the limit of specific training for different kernels and implements
blind debluring.
\begin{figure}[t]
    \begin{center}
    \includegraphics[width=1\linewidth]{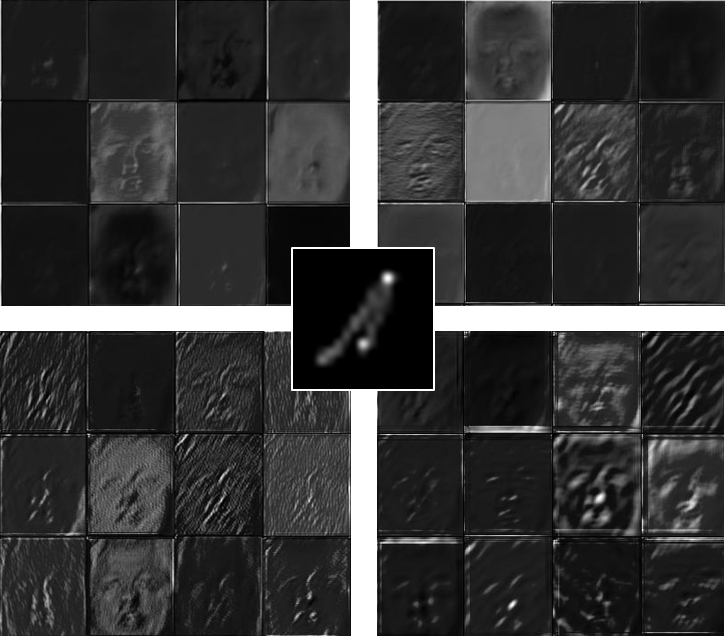}
    \end{center}
    \caption{Responses of different convolution kernels in the first inception stage. 
    Kernel size in left-top: $3\times3$, right-top: $5\times5$, left-bottom: $7\times7$, 
    right-bottom: $14\times14$. Middle: blur kernel shape. Responses show similar 
    characteristic to the blur kernel.}
    \label{fig:response}
\end{figure}
\subsection{Smoothness and Facial Regularizations}
One major contribution of our method is to restore face images as well keeping individual's identification and 
structure information. To achieve this goal, multiple regularizations or losses are used for model training 
which is illustrated in Figure~\ref{fig:pipeline}.
First, we adopt L2 regularization between result image restored by the network and its
ground-truth sharp image using 
\begin{equation}
    \ell_{L2} = \frac{1}{WHC}\sum_{i,j,l}\|\hat{x}_{i,j,l} - x_{i,j,l}\|^2
\end{equation}
where $W$, $H$ are width and height of an image, $C$ is channel number, and $i,j,l$
denote indexes along these three dimensions.
L2 regularization computes distance between two images in
Eulerian space and it is the most common loss function for training. It provides gurantee for
training process to reach roughly convergence. However, L2 
regularization just neglects texture details and information for recognition. 
Also, there are unrealistic noisy region in the generated image after deep neural network.
TV loss(Total Variation loss) in \cite{Kaur2017Photo} is adopted to resolve this 
problem. TV loss measures smoothness of an image which can avoid abrupt changes and
eliminate artifacts. It is written as
\begin{multline}
    \ell_{TV} = \sum_{i,j,k}((\hat{x}_{i,j+1,l} - \hat{x}_{i,j,l})^2 \\
     + (\hat{x}_{i+1,j,l}-\hat{x}_{i,j,l})^2)
\end{multline}
TV loss can be seen as the sum of local gradient squared norm between pixels.
It is also a special case of ``sparse prior'' in \cite{Son2017Fast}. It encourages training to converge to smooth images instead
of local minimum of L2 loss function.
\par
To keep structure information during restoration, we then propose a facial loss for facial 
identity preserving. Content loss in \cite{Ledig2016Photo} or
facial semantic structure loss in \cite{Kaur2017Photo} both use another deep network to
extract presentative features directly from image data. For better performance verification tasks, 
we extract fully connected layer feature of our pre-trained  
CNN model based on ResNet trained for face recognition instead of using feature 
map in convolution layer of VGG. The last fully connected layer has much less parameters than
convolution layers so it avoids introducing more computations.
Thus, facial loss is defined as
the L2 distance between extracted feaures of ground-truth and restored image:
\begin{equation}
    \ell_{face} = \|\Phi(\hat{x})-\Phi(x)\|^2
\end{equation}
where $\Phi$ denotes the final fully connected layer of face recognition model. 
Then, the total loss function for training is formulated as the weighted combination of L2-loss, 
TV loss and facial loss:
\begin{equation}
    L = \ell_{L2} + \alpha\ell_{TV} + \beta\ell_{face}
    \label{equa:loss}
\end{equation}
where $\alpha$ and $\beta$ are balancing factors when L2-loss weight is
fixed to 1. The total loss function regulates the training into convergence
while maintaining facial information and reducing noise during restoration. More trainning details
and analyses are in section ~\ref{Training} and ~\ref{Analysis}.

\begin{figure*}[t]
    \centering
    \subfigure[blurry]{\label{fig:4-3}
        \begin{minipage}[b]{0.13\linewidth}
            \includegraphics[width=1\linewidth]{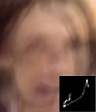}\\
            \includegraphics[width=1\linewidth]{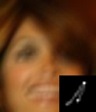}
        \end{minipage}}
    \subfigure[Levin \etal]{\label{fig:4-4}
    \begin{minipage}[b]{0.13\linewidth}
    \includegraphics[width=1\linewidth]{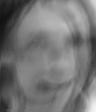}\\
    \includegraphics[width=1\linewidth]{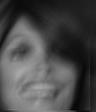}
    \end{minipage}}
    \subfigure[Sun \etal]{\label{fig:4-5}
    \begin{minipage}[b]{0.13\linewidth}
    \includegraphics[width=1\linewidth]{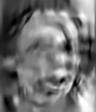}\\
    \includegraphics[width=1\linewidth]{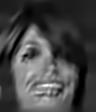}
    \end{minipage}}
    \subfigure[Pan \etal]{\label{fig:4-6}
    \begin{minipage}[b]{0.13\linewidth}
    \includegraphics[width=1\linewidth]{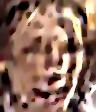}\\
    \includegraphics[width=1\linewidth]{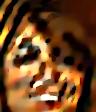}
    \end{minipage}}
    \subfigure[Perrone \etal]{\label{fig:4-7}
    \begin{minipage}[b]{0.13\linewidth}
    \includegraphics[width=1\linewidth]{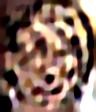}\\
    \includegraphics[width=1\linewidth]{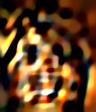}
    \end{minipage}}
    \subfigure[Krishnan \etal]{\label{fig:4-8}
    \begin{minipage}[b]{0.13\linewidth}
    \includegraphics[width=1\linewidth]{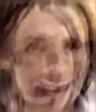}\\
    \includegraphics[width=1\linewidth]{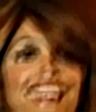}
    \end{minipage}}
    \subfigure[ours]{\label{fig:4-9}
    \begin{minipage}[b]{0.13\linewidth}
    \includegraphics[width=1\linewidth]{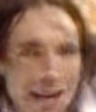}\\
    \includegraphics[width=1\linewidth]{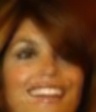}
    \end{minipage}}
    \caption{Comparision of restored images between several state-of-the-art methods. While many algorithms
     come into collapses, our method still restore distinguishable face images.}
    \label{fig:qual results}
\end{figure*}
\subsection{CNN Architecture}
Restoring network in Figure~\ref{fig:pipeline} is designed as ResNet \cite{He2015Deep} structure with Inception\cite{Szegedy2014Going} 
modules. 
In order to handle those unknown blurry images in all different blur kernels,
multi-scale convolution is adopted to extract features including both blurred edge
details and long distance movements. Each inception block in Figure~\ref{fig:CNN} has 
five convolution kernel scales from size 1 to size 14. The larger kernels strengthen
lines-like response in larger scale while the smaller kernels extract texture
details. As shown in Figure~\ref{fig:response}, $3\times3$ and $5\times5$ kernels
focus more on texture features than $7\times7$ kernels, and responses of $14\times14$ kernels are
more abstract. Also, responses of the blurry image under multi-scale convolutions show
similar shapes corresponding to the blur kernel. It indicates that our network can
distinguish implicit blur kernel shape in a single blurry image. This ability of 
feature representation benefits deblurred image restoration by selecting and non-linear
combining of feature maps. 
\par
Multi-scale structure enlarges network's width that normally can improve 
model's performance\cite{Szegedy2014Going}. However, computation complexity rises 
rapidly when 
network's depth and width increase. In every inception block we use depth-wise 
convolution to reduce channel dimension to half of it before each standard expensive
convolution layer. Then, outputs of different convolution size are concatenated in channel
dimension and reduced again. Additional improvement is analysed in section~\ref{Analysis}.
\par
As shown
in Figure~\ref{fig:CNN}, this network has six Inception modules which have 
multi-scale convolution layers respectively. Each module's input is added to its
output and modules are stacked upon each other. First convolution layer out of Inception 
modules has $3\times3$ convolution kernels and outputs 64 channels. 
Final convolution layer reduces the output to three channels with $1\times1$ convolution 
and gets the restoring results. We use leaky ReLU in \cite{maas2013rectifier} 
as the nonlinear activation function following convolution layers. 
\par
Thus, our designed architecture with multi-scale convolutions makes it possible 
for restoring model to deal with complicated inputs and enhance adaptability of various
blur scales. Meanwhile, reduction using $1\times1$ convolution kernel avoids explosive growth 
of model parameters. This design enables training process to be controllable and 
reduces computational complexity.


\section{Experiments}
\subsection{Training}\label{Training}
We use CASIA WebFace Database \cite{Yi2014Learning} as training sets. 
It has 494,414 images from 
10575 diffrent subjects. All images are aligned and cropped into $112\times96$ 
by landmarks. The network is trained with pairs of sharp image and its blurry
image synthetized with kernel describe in section ~\ref{kernel_gen}. Valid kernel size
is randomly chosen and the mass center of kernel is shifted to the middle. \par
We adopt RMSProp optimizer and apply exponential decay to the learning rate.
At the begining, learning rate is set to 0.001 and it decays along with steps
increasing. We keep learning rate small and train for a long time to prevent 
divergence and gradient exploding. When loss terms stop decreasing
the learning rate is cut to half of it. 
Balancing weights between TV and facial loss terms are adjusted manually
during different stages of training process. The training is stopped
when restored latent images look real and sharp as \cite{Shrivastava2016Learning} do.
\subsection{Implentation Details}
We implement our method and perform experiments with various images and blur kernels
on Intel Xeon E5 CPU and NVIDIA GeForce GTX 1080 GPU. Time consuming and restored 
results of our method are compared with several other methods \cite{Krishnan2011Blind,Levin2011Efficient,Pan2014Deblurring,Perrone2014Total,Michaeli2014Blind,Chakrabarti2016A}  . 
Those methods' MATLAB implentations are published online by their authors. Our model is
trained and tested with Tensorflow. Since our method is one-step 
that estimating sharp images from blurry inputs, test stage is a simple network forwarding
process.\par
\begin{figure*}[ht]
    \centering
    \subfigure[]{
    \begin{minipage}[b]{0.11\linewidth}
        \includegraphics[width=1\linewidth]{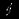}\\
        \includegraphics[width=1\linewidth]{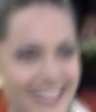}\\
        \includegraphics[width=1\linewidth]{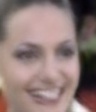}
    \end{minipage}}
    \subfigure[]{
        \begin{minipage}[b]{0.11\linewidth}
            \includegraphics[width=1\linewidth]{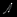}\\
            \includegraphics[width=1\linewidth]{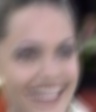}\\
            \includegraphics[width=1\linewidth]{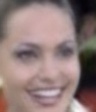}
        \end{minipage}}
    \subfigure[]{
    \begin{minipage}[b]{0.11\linewidth}
        \includegraphics[width=1\linewidth]{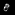}\\
        \includegraphics[width=1\linewidth]{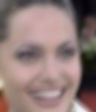}\\
        \includegraphics[width=1\linewidth]{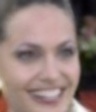}
    \end{minipage}}
    \subfigure[]{
    \begin{minipage}[b]{0.11\linewidth}
        \includegraphics[width=1\linewidth]{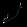}\\
        \includegraphics[width=1\linewidth]{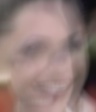}\\
        \includegraphics[width=1\linewidth]{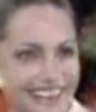}
    \end{minipage}}
    \subfigure[]{
    \begin{minipage}[b]{0.11\linewidth}
        \includegraphics[width=1\linewidth]{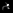}\\
        \includegraphics[width=1\linewidth]{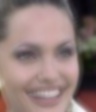}\\
        \includegraphics[width=1\linewidth]{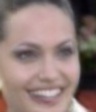}
    \end{minipage}}
    \subfigure[]{
    \begin{minipage}[b]{0.11\linewidth}
        \includegraphics[width=1\linewidth]{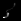}\\
        \includegraphics[width=1\linewidth]{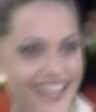}\\
        \includegraphics[width=1\linewidth]{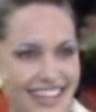}
    \end{minipage}}
    \subfigure[]{
    \begin{minipage}[b]{0.11\linewidth}
        \includegraphics[width=1\linewidth]{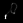}\\
        \includegraphics[width=1\linewidth]{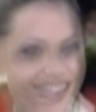}\\
        \includegraphics[width=1\linewidth]{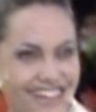}
    \end{minipage}}
    \subfigure[]{
    \begin{minipage}[b]{0.11\linewidth}
        \includegraphics[width=1\linewidth]{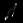}\\
        \includegraphics[width=1\linewidth]{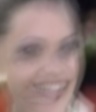}\\
        \includegraphics[width=1\linewidth]{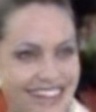}
    \end{minipage}}
    \caption{Our restored results of different benchmark kernels. Top row: blur kernels, middle
    row: corresponding blurry inputs, bottom row: restored outputs. Blurry face images
    in different blur shapes and sizes are restored with satisfactory results.}
    \label{fig:qual results2}
\end{figure*}

Because our method aims at restoring blurry face images, experiments for the evaluation 
are performed on 
widely used benchmark face datasets such as LFW \cite{LFWTech} and FaceScrub \cite{Ng2015A}.
All test images are aligned and cropped in the same way as training.
Then, those images are convolved with eight blur kernels of different sizes from \cite{Levin2011Efficient}
and are shown in the
first row of Figure~\ref{fig:qual results2} . We also test our method
on linear motion kernel.
All the results deblurred by different methods are saved for quantitative and qualitative comparision.

\subsection{Time Consuming Comparision}
Tabel~\ref{table:time} shows the averaged time consuming of resotring a single face image
in size $112\times96$. Those methods are implented with their default settings and 
parameters. Methods 
which use EPLL as the non-blind 
deconvolution step\cite{patchdeblur_iccp2013, Chakrabarti2016A, Michaeli2014Blind} 
are implemented with the same EPLL code package. Although training deep
neural network needs a large amount of time, inference time during test is independent
with training. Back-propagation of loss function during training does not affect inference time.
Without itertations between latent image and estimated blur kernel, our method 
runs very fast in the test stage using GPU. Our method outperforms the fastest algorithm
more than 100 times shown in Tabel~\ref{table:time}. 
\begin{table*}
    \begin{center}
    \begin{tabular}{|l|c|c|c|c|}
    \hline
    Method & Sun \etal \cite{patchdeblur_iccp2013} & Chakrabarti \cite{Chakrabarti2016A} & Krishnan \etal\cite{Krishnan2011Blind} & Levin \etal\cite{Levin2011Efficient} \\
    \hline
    Mean Time(s) & 89.06 & 20.51 & 1.84 & 22.74  \\
    \hline
    Method &Michaeli \etal \cite{Michaeli2014Blind} & Pan \etal \cite{Pan2014Deblurring} & Perrone \etal\cite{Perrone2014Total} & ours \\
    \hline
    Mean Time(s) & 417.32 & 10.27 & 18.36 & \textbf{0.0148} \\
    \hline
    \end{tabular}
    \end{center}
    \caption{Time consuming comparision of a single image restoration. We take the averaged time of eight types 
        blurry datasets. Our method runs very fast.}
    \label{table:time}    
\end{table*}

\begin{table*}[ht]
    \begin{center}
    \begin{tabularx}{\textwidth}{p{2.3cm}*{9}{>{\centering\arraybackslash}X}}
    \toprule
            & kernel 1 & kernel 2 &kernel 3 &kernel 4 &kernel 5 &kernel 6 &kernel 7 &kernel 8 & mean \\
    \midrule
    Krishnan \etal & 92\% & 88\% & 92\% & 82\% & 92\% & 86\% & 96\% & 78\% & 88.25\% \\
    Pan \etal & 86\% & 80\% & 84\% & 64\% & 84\% & 66\% & 72\% & 72\% & 76\% \\
    Levin \etal & 92\% & 80\% & 96\% & 48\% & \textbf{98}\% & 74\% & 70\% & 62\% & 77.5\% \\
    Perr \etal & 46\% & 52\% & 46\% & 48\% & 62\% & 48\% & 60\% & 66\% & 53.5\% \\
    Sun \etal & 86\% & 88\% & 94\% & 72\% & 90\% & 80\% & 86\% & 78\% & 84.25\% \\
    Ours & \textbf{98}\% & \textbf{98}\% & \textbf{98}\% & \textbf{96}\% &
    \textbf{98}\% & \textbf{98}\% & \textbf{98}\% & \textbf{98}\% & \textbf{97.75}\%\\  
    \bottomrule
    \end{tabularx}
    \end{center}
    \caption{Verification accuracy of restored face images on subset of LFW.
    Experiments on eight benchmark kernels in \cite{Levin2011Efficient} are all illustrated.
    Kernels are sorted according to order in Figure~\ref{fig:qual results2}.
    Best results are in bold.}
    \label{table:acc}
\end{table*}
\begin{figure}[hb]
    \centering
    \includegraphics[width=1\linewidth]{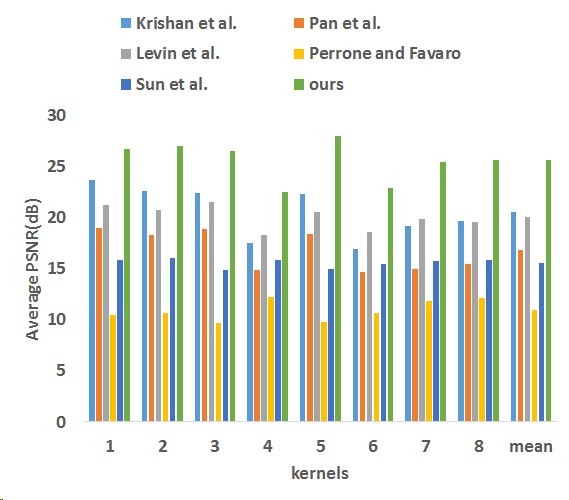}
    \caption{Results of PSNR on LFW. Kernels are sorted according to order in Figure~\ref{fig:qual results2}.}
    \label{fig:PSNR}
\end{figure}

\subsection{Qualitative Results}
We test five methods in Tabel~\ref{table:time} which can obtain meaningful deblurred 
face images on eight type blurry test datasets and compare them with ours. A restored example
are shown in Figure~\ref{fig:qual results}. The blurry input in
subfigure~\ref{fig:4-3} is sythesized with the right-bottom kernel from original image.
~\ref{fig:4-4} and~\ref{fig:4-5} are in grayscale because Levin \etal \cite{Levin2011Efficient} and
Sun \etal \cite{patchdeblur_iccp2013} can only handle one channel of a RGB image.
Except Krishnan \etal \cite{Krishnan2011Blind} in~\ref{fig:4-8} sharpen edges in blurry input to 
some extent, other 
methods all fail to restore a distinguishable human face due to severe blur and small input size.
Although kernel scale is
relatively large to image size($27\times27$ vs. $112\times96$), our method still
performs a fairly good result with a distinct face.\par
Figure~\ref{fig:qual results2} illustrates that our model can handle blur kernels in 
different size and shape. Results in third row show that our model has
generalization capability since all test blur kernels have never appeared in training. 
Compared to blurry degraded images in Figure~\ref{fig:qual results2} second row,
one-step restoring method improves face image quality of visualization significantly.

\subsection{Quantitative Results}\label{Quantitative}
Our method is evaluated in terms of recognition rate and PSNR on face datasets and
benchmark of blur kernels in \cite{Levin2011Efficient}. Because traditional methods 
consume too much 
time(Table~\ref{table:time}), we have to choose 50 pairs randomly from 3000 test
pairs of LFW and compute quantitative results on this small size datasets. Performance
comparision is illustrated in Figure~\ref{fig:PSNR}. Krishnan \etal \cite{Krishnan2011Blind} and 
Levin \etal \cite{Levin2011Efficient} show better results than Sun \etal \cite{patchdeblur_iccp2013}, 
Perrone \etal \cite{Perrone2014Total} and Pan \etal \cite{Pan2014Deblurring}. Performance 
of our method falls 
a little when handling large blur kernel size, but it's still beyond others a lot.
\par
Restored face images should maintain their identity information which is useful for
face recognition. For the chosen test pairs of LFW, one of each pair keeps its 
orginal image while the other is restored from the blurry image. Blurry images are also
synthetized by eight kernel in \cite{Levin2011Efficient}. We apply a face verification
model trained on MS-Celeb-1M \cite{guo2016msceleb}
to test the verification accuracy of all face image pairs restored from different 
methods. Table~\ref{table:acc} shows verification accuracy of all restoration methods.
By reconstructing facial structure and information, our model achieves the best
accuracy performance. We also test our model on all images of LFW thanks to very 
fast running speed and show the results in Table~\ref{table:acc2}. It shows that 
blurry images affect verification accuracy badly especially when blur scale is large. 
Our method can eliminate this influence obviously.
\par
Tabel~\ref{table:acc} and Table~\ref{table:acc2} show that blur disturbance in large
scale may reduce performance of verification model sharply while small size blurs have
much slighter influence. Our restoring method can benefit face verification no matter 
how the blur situation changes. \par
PSNR of linear motion deblurred results is also illustrated
in Table~\ref{table:linear}. Linear kernel has length of 15 pixels and angle of $45^{\circ}$.
We compare ours with the superior one \cite{Krishnan2011Blind} of other methods and compute PSNR.
Although linear motion kernels never appeared in our training, our model also outputs
plausible results.
\par
\begin{figure}
    \centering
    \includegraphics[width=1\linewidth]{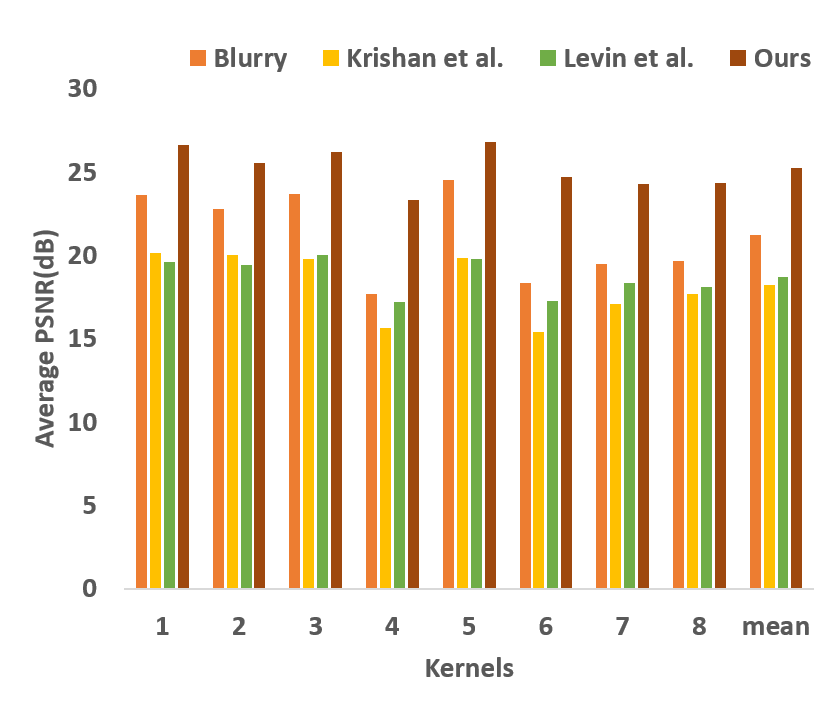}
    \caption{PSNR comparision on FaceScrub. Kernels are sorted according to order in Figure~\ref{fig:qual results2}.
    Our results are still comparable to state-of-the-art methods.}
    \label{fig:PSNR_FS}
\end{figure}

We also evaluate our method on FaceScrub and compare it with 
\cite{Levin2011Efficient,Krishnan2011Blind} shown in Figure~\ref{fig:PSNR_FS}.
Our method obtain better restoration performances for both different face images and various
blur kernels.
Comprehensive experiments demonstrate that our method can improve images quality cross face
datasets and blur kernels.

\begin{table}
    \begin{center}
    \begin{tabularx}{\linewidth}{*{5}{>{\centering\arraybackslash}X}}
        \toprule
                & kernel 1 & kernel 2 & kernel 3 & kernel 4\\
        \midrule
        Blurry & 91.48\% & 86.98\% & 97.45\% & 65.63\% \\
        Ours & 96.70\% & 97.48\% & 98.38\% & 95.00\%\\
        \bottomrule
    \end{tabularx}
    
    \begin{tabularx}{\linewidth}{*{5}{>{\centering\arraybackslash}X}}
                & kernel 5 & kernel 6 & kernel 7 & kernel 8\\
        \midrule
        Blurry & 98.07\% & 93.78\% & 87.97\% & 81.18\%\\
        Ours & 98.28\% & 96.98\% & 96.95\% & 96.47\%\\
        \bottomrule
    \end{tabularx}
    \end{center}
\caption{Veification accuracy comparision between blurry images and our restored results
            on the whole LFW. 
            Our verification model has 98.98\% accuracy on the standard LFW benchmark.}
\label{table:acc2}
    
\end{table}



\begin{table}
    \begin{center}
    \begin{tabularx}{\linewidth}{*{4}{>{\centering\arraybackslash}X}}
        \toprule
                & Blurry & Kris & Ours \\
        \midrule
        PSNR(dB) & 24.95 & 22.83 & \textbf{27.38} \\
        \bottomrule
    \end{tabularx}
    \end{center}
    \caption{Restored results of linear motion kernel blur which never appeared in training.
    L = $15$, angle = $45^{\circ}$.
    }
    \label{table:linear}
\end{table}

\subsection{Analysis on Network and Regularization}\label{Analysis}
In each inception module, depth-wise convolution is adopted to reduce depth dimension which 
saves nearly $40\%$ of model size. 
A new model is built without $1\times1$ convolution reduction and following kernels
are modified to $k\times k\times64\times32$, $k$ is each kernel's size. The modified network
is trained with the same iterations and tested on LFW database. During training we found
that modified model reaches to convergence slower than original model. Tabel~\ref{table:depth} shows
that depth-wise convolution layers also enable deep network to achieve a better performance
besides reduce computational complexity. 
\par
\begin{figure}
    \centering
    \subfigure[]{
    \includegraphics[width=0.3\linewidth]{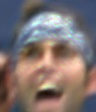}}
    \subfigure[]{
    \includegraphics[width=0.3\linewidth]{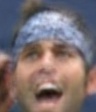}}
    \subfigure[]{
    \includegraphics[width=0.3\linewidth]{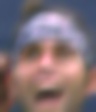}}
    \caption{Restored results with different regularization weights.
    left: $\alpha/\beta$ = 0.2, middle: $\alpha/\beta$ = 1, right:
     $\alpha/\beta$ = 5. Proper weights ratio is critical to trainging process.}
    \label{fig:regu}
\end{figure}

Regularization terms in Eq.~\ref{equa:loss} help the network to avoid local minimum and overfitting.
The ratio between facial loss and 
TV loss decides quality of restored images. Figure~\ref{fig:regu} shows different results restored
of models trained by different regularization weights. When TV loss contributes much less than facial loss,
there are noisy artifacts in results. On the other hand, increasing 
$\alpha/\beta$ to 5 suppresses texture features and leads to over-smoothness.
Balancing regularization terms during training
plays an important role of the implentation. 

\begin{table}
    \begin{center}
        \begin{tabular}{ccc}
            \toprule
                    & PSNR & Verification Acc.(\%) \\
            \midrule
            Our model & 25.63 &  97.75\\
            without depth-wise conv & 23.05 & 88.5 \\
            \bottomrule
        \end{tabular}
    \end{center}
    \caption{Comparision between models with/without depth-wise convolution layers.}
    \label{table:depth}
\end{table}

\section{Conclusion}
In this paper,  a very fast and effective one-step restoring method for 
blurry face images is proposed. We design a CNN to handle various blur kernels without any priors
or estimations. Proposed method breaks the training limit of deep learning method that 
can only handle specific blurs. Although face images are lack of salient edges and relatively
small, our restoration model improves image quality significantly and keeps facial identity
by adding smoothness and facial regularizations. Our model runs at high speed which is
compatible in recognition applications. Going deep into real blurry images will be interesting
future work.

{\small
\bibliographystyle{ieee}
\bibliography{deblur}
}

\section{More Experiments Results}
\begin{figure*}[t]
    \centering
    \subfigure[blurry]{
    \begin{minipage}[b]{0.13\linewidth}
        \includegraphics[width=1\linewidth]{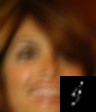}\\
        \includegraphics[width=1\linewidth]{Cindy_k2//Cindy_blurry.jpg}\\
        \includegraphics[width=1\linewidth]{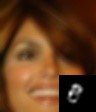}\\
        \includegraphics[width=1\linewidth]{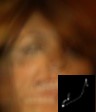}\\
        \includegraphics[width=1\linewidth]{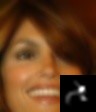}\\
        \includegraphics[width=1\linewidth]{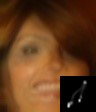}\\
        \includegraphics[width=1\linewidth]{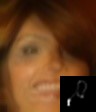}\\
        \includegraphics[width=1\linewidth]{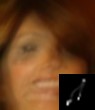}
    \end{minipage}}
    \subfigure[Levin \etal]{
    \begin{minipage}[b]{0.13\linewidth}
    \includegraphics[width=1\linewidth]{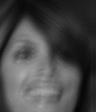}\\
    \includegraphics[width=1\linewidth]{Cindy_k2//Cindy_Levin.jpg}\\
    \includegraphics[width=1\linewidth]{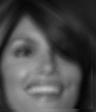}\\
    \includegraphics[width=1\linewidth]{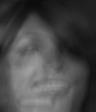}\\
    \includegraphics[width=1\linewidth]{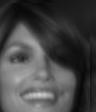}\\
    \includegraphics[width=1\linewidth]{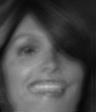}\\
    \includegraphics[width=1\linewidth]{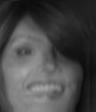}\\
    \includegraphics[width=1\linewidth]{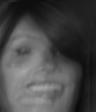}
    \end{minipage}}
    \subfigure[Sun \etal]{
    \begin{minipage}[b]{0.13\linewidth}
    \includegraphics[width=1\linewidth]{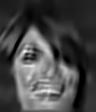}\\
    \includegraphics[width=1\linewidth]{Cindy_k2//Cindy_Sun.jpg}\\
    \includegraphics[width=1\linewidth]{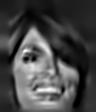}\\
    \includegraphics[width=1\linewidth]{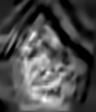}\\
    \includegraphics[width=1\linewidth]{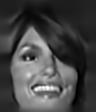}\\
    \includegraphics[width=1\linewidth]{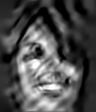}\\
    \includegraphics[width=1\linewidth]{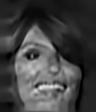}\\
    \includegraphics[width=1\linewidth]{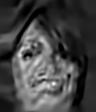}
    \end{minipage}}
    \subfigure[Pan \etal]{
    \begin{minipage}[b]{0.13\linewidth}
    \includegraphics[width=1\linewidth]{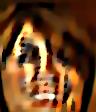}\\
    \includegraphics[width=1\linewidth]{Cindy_k2//Cindy_Pan.jpg}\\
    \includegraphics[width=1\linewidth]{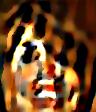}\\
    \includegraphics[width=1\linewidth]{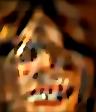}\\
    \includegraphics[width=1\linewidth]{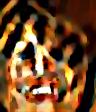}\\
    \includegraphics[width=1\linewidth]{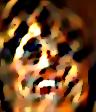}\\
    \includegraphics[width=1\linewidth]{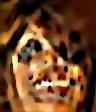}\\
    \includegraphics[width=1\linewidth]{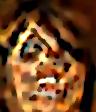}
    \end{minipage}}
    \subfigure[Perrone \etal]{
    \begin{minipage}[b]{0.13\linewidth}
    \includegraphics[width=1\linewidth]{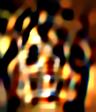}\\
    \includegraphics[width=1\linewidth]{Cindy_k2//Cindy_Perr.jpg}\\
    \includegraphics[width=1\linewidth]{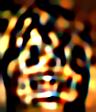}\\
    \includegraphics[width=1\linewidth]{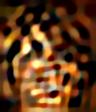}\\
    \includegraphics[width=1\linewidth]{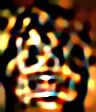}\\
    \includegraphics[width=1\linewidth]{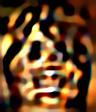}\\
    \includegraphics[width=1\linewidth]{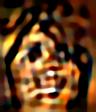}\\
    \includegraphics[width=1\linewidth]{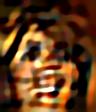}
    \end{minipage}}
    \subfigure[Krishnan \etal]{
    \begin{minipage}[b]{0.13\linewidth}
    \includegraphics[width=1\linewidth]{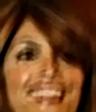}\\
    \includegraphics[width=1\linewidth]{Cindy_k2//Cindy_Kris.jpg}\\
    \includegraphics[width=1\linewidth]{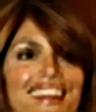}\\
    \includegraphics[width=1\linewidth]{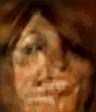}\\
    \includegraphics[width=1\linewidth]{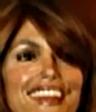}\\
    \includegraphics[width=1\linewidth]{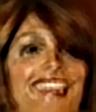}\\
    \includegraphics[width=1\linewidth]{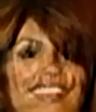}\\
    \includegraphics[width=1\linewidth]{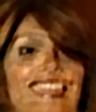}
    \end{minipage}}
    \subfigure[ours]{
    \begin{minipage}[b]{0.13\linewidth}
    \includegraphics[width=1\linewidth]{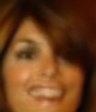}\\
    \includegraphics[width=1\linewidth]{Cindy_k2//Cindy_rec.jpg}\\
    \includegraphics[width=1\linewidth]{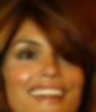}\\
    \includegraphics[width=1\linewidth]{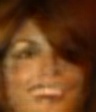}\\
    \includegraphics[width=1\linewidth]{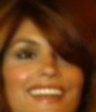}\\
    \includegraphics[width=1\linewidth]{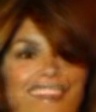}\\
    \includegraphics[width=1\linewidth]{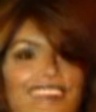}\\
    \includegraphics[width=1\linewidth]{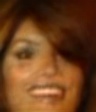}
    \end{minipage}}
    \caption{Comparision of restored results on LFW.}
    \label{fig:LFW}
\end{figure*}

\begin{figure*}[t]
    \centering
    \subfigure[kernel 1]{
    \begin{minipage}[b]{0.11\linewidth}
        \includegraphics[width=1\linewidth]{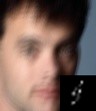}\\
        \includegraphics[width=1\linewidth]{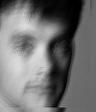}\\
        \includegraphics[width=1\linewidth]{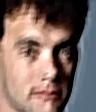}\\
        \includegraphics[width=1\linewidth]{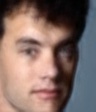}
    \end{minipage}}
    \subfigure[kernel 2]{
    \begin{minipage}[b]{0.11\linewidth}
        \includegraphics[width=1\linewidth]{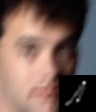}\\
        \includegraphics[width=1\linewidth]{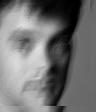}\\
        \includegraphics[width=1\linewidth]{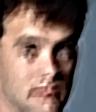}\\
        \includegraphics[width=1\linewidth]{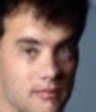}
    \end{minipage}}
    \subfigure[kernel 3]{
    \begin{minipage}[b]{0.11\linewidth}
        \includegraphics[width=1\linewidth]{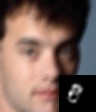}\\
        \includegraphics[width=1\linewidth]{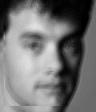}\\
        \includegraphics[width=1\linewidth]{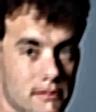}\\
        \includegraphics[width=1\linewidth]{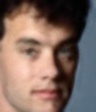}
    \end{minipage}}
    \subfigure[kernle 4]{
    \begin{minipage}[b]{0.11\linewidth}
        \includegraphics[width=1\linewidth]{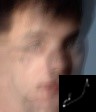}\\
        \includegraphics[width=1\linewidth]{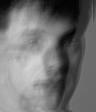}\\
        \includegraphics[width=1\linewidth]{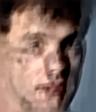}\\
        \includegraphics[width=1\linewidth]{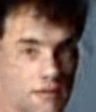}
    \end{minipage}}
    \subfigure[kernel 5]{
    \begin{minipage}[b]{0.11\linewidth}
        \includegraphics[width=1\linewidth]{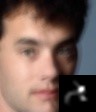}\\
        \includegraphics[width=1\linewidth]{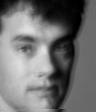}\\
        \includegraphics[width=1\linewidth]{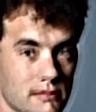}\\
        \includegraphics[width=1\linewidth]{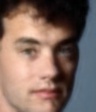}
    \end{minipage}}
    \subfigure[kernel 6]{
    \begin{minipage}[b]{0.11\linewidth}
        \includegraphics[width=1\linewidth]{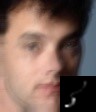}\\
        \includegraphics[width=1\linewidth]{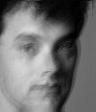}\\
        \includegraphics[width=1\linewidth]{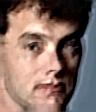}\\
        \includegraphics[width=1\linewidth]{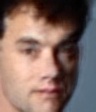}
    \end{minipage}}
    \subfigure[kernel 7]{
    \begin{minipage}[b]{0.11\linewidth}
        \includegraphics[width=1\linewidth]{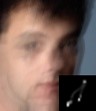}\\
        \includegraphics[width=1\linewidth]{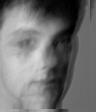}\\
        \includegraphics[width=1\linewidth]{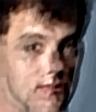}\\
        \includegraphics[width=1\linewidth]{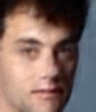}
    \end{minipage}}
    \subfigure[kernel 8]{
        \begin{minipage}[b]{0.11\linewidth}
            \includegraphics[width=1\linewidth]{FS//Tom_k8//blurry.jpg}\\
            \includegraphics[width=1\linewidth]{FS//Tom_k8//Tom_Hanks_Levin.jpg}\\
            \includegraphics[width=1\linewidth]{FS//Tom_k8//Tom_Hanks_Kris.jpg}\\
            \includegraphics[width=1\linewidth]{FS//Tom_k8//Tom_Hanks_rec.jpg}
        \end{minipage}}
    \caption{Restored images on FaceScrub. Compared with Levin \etal and Krishnan \etal.
    First row: Blurry face images generated by different kernels. Second row: Results 
    restored by Levin \etal. Third row: Results restored by Krishnan \etal. Last row:
    Our restored results.}
    \label{fig:fs}
\end{figure*}

In this section, more detailed qualitative results are shown to evaluate the effectiveness
of the proposed method.
Comparison of resored results on LFW between six different methods including ours is shown 
in Figure~\ref{fig:LFW}. Figure~\ref{fig:fs} illustrates resored images on FaceScrub as well.
Proposed method can handle various blur kernels and is adaptable at different face images.

\end{document}